\documentclass{edm_template}
\usepackage{xcolor}
\usepackage{graphicx}

\begin{document}

\newcommand{\ka}{{\sc Khan Academy}}
\newcommand{\assistments}{{\sc Assistments}}
\newcommand{\synthetic}{{\sc Synthetic}}
\newcommand{\spanish}{{\sc Spanish}}
\newcommand{\statics}{{\sc Statics}}

\title{How Deep is Knowledge Tracing?}

%
%
%
%
%

\numberofauthors{3} 
\author{
\alignauthor
Mohammad Khajah\\
       \affaddr{Dept. of Computer Science}\\
       \affaddr{University of Colorado}\\
       \affaddr{Boulder, Colorado 80309}\\
       \email{\normalsize \tt mohammad.khajah@colorado.edu}
\alignauthor
Robert V. Lindsey\\
       \affaddr{Dept. of Computer Science}\\
       \affaddr{University of Colorado}\\
       \affaddr{Boulder, Colorado 80309}\\
       \email{\normalsize \tt robert.lindsey@colorado.edu}
\alignauthor 
Michael C. Mozer\\
       \affaddr{Dept. of Computer Science}\\
       \affaddr{University of Colorado}\\
       \affaddr{Boulder, Colorado 80309}\\
       \email{\normalsize \tt mozer@colorado.edu}
}
\date{30 July 1999}

\maketitle
\begin{abstract}
In theoretical cognitive science, there is a tension between highly structured
models whose parameters have a direct psychological interpretation and highly
complex, general-purpose models whose parameters and representations are
difficult to interpret.  The former typically provide more insight into
cognition but the latter often perform better. This tension has recently
surfaced in the realm of educational data mining, where a deep learning
approach to predicting students' performance as they work through a series of
exercises---termed \textit{deep knowledge tracing} or \textit{DKT}---has
demonstrated a stunning performance advantage over the mainstay of the field,
\textit{Bayesian knowledge tracing} or \textit{BKT}.  In this article, we
attempt to understand the basis for DKT's advantage by considering the sources
of statistical regularity in the data that DKT can leverage but which BKT
cannot. We hypothesize four forms of regularity that BKT fails to exploit:
recency effects, the contextualized trial sequence, inter-skill similarity, and
individual variation in ability.  We demonstrate that when BKT is extended to
allow it more flexibility in modeling statistical regularities---using
extensions previously proposed in the literature---BKT achieves a level of
performance indistinguishable from that of DKT. We argue that while DKT is a
powerful, useful, general-purpose framework for modeling student learning, its
gains do not come from the discovery of novel representations---the fundamental
advantage of deep learning.  To answer the question posed in our title,
knowledge tracing may be a domain that does \textit{not} require `depth';
shallow models like BKT can perform just as well and offer us greater
interpretability and explanatory power.

%


\section{INTRODUCTION}
In the past forty years, machine learning and cognitive science have undergone
many paradigm shifts, but few have been as dramatic as the recent surge of
interest in \textit{deep learning} \cite{leCunBengioHinton2015}.  Although deep
learning is little more than a re-branding of neural network techniques popular
around 1990, deep learning has achieved some remarkable results thanks to much
faster computing resources and much larger data sets than were available in
1990. Deep learning underlies state-of-the-art systems in speech recognition,
language processing, and image classification
\cite{leCunBengioHinton2015,Schmidhuber2015}.  Deep learning also is responsible
for systems that can produce captions for images \cite{imagecaptions}, create
synthetic images \cite{imagegeneration}, play video games \cite{DeepMind2015} and
even Go \cite{DeepMindGo2016}.

The `deep' in deep learning refers to multiple levels of representation
transformation that lie between model inputs and outputs. For example, 
an image-classification model may take pixel values as input and produce a 
labeling of the objects in the image as output. Between the input and output 
is a series of representation transformations that construct successively
higher-order features---features that are less sensitive to lighting
conditions and the position of objects in the image, and more sensitive
to the identities of the objects and their qualitative relationships.
The features discovered by deep learning exhibit a complexity and subtlety
that make them difficult to analyze and understand
(e.g., \cite{ZeilerFergus2013}). Furthermore, no human engineer could
wire up a solution as thorough and accurate as solutions discovered by
deep learning. Deep learning models are fundamentally \textit{nonparametric},
in the sense that interpreting individual weights and individual unit
activations in a network is pretty much impossible. This opacity
is in stark contrast to parametric models, e.g., linear regression,
where each of the coefficients has a clear interpretation in terms of
the problem at hand and the input features.

In one domain after the next, deep learning has achieved gains over
traditional approaches.  Deep learning discards hand-crafted features 
in favor of representation learning, and deep learning often ignores
domain knowledge and structure in favor of massive data sets and general 
architectural constraints on models (e.g., models with spatial locality to 
process images, and models with local temporal constraints to process time 
series).

It was inevitable that deep learning would be applied to student-learning data
\cite{Piechetal2015}. This domain has traditionally been the purview of the
educational data mining community, where \textit{Bayesian knowledge tracing},
or \textit{BKT}, is the dominant computational approach \cite{corbett1995}.
The deep learning approach to modeling student data, termed \textit{deep
knowledge tracing} or \textit{DKT}, created a buzz when it appeared at the
Neural Information Processing Systems Conference in December 2015, including
press inquiries (N. Heffernan, personal communication) and descriptions of the
work in the blogosphere (e.g., \cite{Golden2016}). Piech et al.\
\cite{Piechetal2015} reported substantial improvements in prediction
performance with DKT over BKT on two real-world data sets (\assistments, \ka) 
and one synthetic data set which was generated under assumptions that
are not tailored to either DKT or BKT. DKT achieves a
reported 25\% gain in AUC (a measure of prediction quality) over the best
previous result on the \assistments\ benchmark.

In this article, we explore the success of DKT.  One approach to this
exploration might be to experiment with DKT, removing components of the model
or modifying the input data to determine which model components and data
characteristics are essential to DKT's performance.  We pursue an alternative
approach in which we first formulate hypotheses concerning the signals in the
data that DKT is able to exploit but that BKT is not.  Given these hypotheses,
we propose extensions to BKT which provide it with additional flexibility,
and we evaluate whether the enhanced BKT can achieve results comparable to DKT.
This procedure leads not only to a better understanding of how BKT and DKT
differ, but also helps us to understand the structure and statistical
regularities in the data source.

\subsection{Modeling Student Learning}
The domain we're concerned with is electronic tutoring systems which employ
cognitive models to track and assess student knowledge.  Beliefs about what a
student knows and doesn't know allow a tutoring system to dynamically adapt its
feedback and instruction to optimize the depth and efficiency of learning.

Ultimately, the measure of learning is how well students are able to apply
skills that they have been taught. Consequently, student modeling is often
formulated as time series prediction: given the series of exercises a student
has attempted previously and the student's success or failure on each exercise, predict
how the student will fare on a new exercise.  Formally, the data consist of a
set of binary random variables indicating whether student $s$ produces
a correct response on trial $t$,
$\{ X_{st} \}$. The data also
include the exercise labels, $\{ Y_{st} \}$, which characterize the exercise.
Secondary data has also been incorporated in models, including the student's
utilization of hints, response time, and characteristics of the specific
exercise and the student's particular history with related exercises \cite{baker2008,
yu2010}.  Although such data improve predictions, the bulk of research in this
area has focused on the primary measure---whether a response is correct or incorrect---and a sensible research
strategy is to determine the best model based on the primary data, and then to
determine how to incorporate secondary data.

The exercise label, $Y_{st}$, might index the specific exercise, e.g., 
$3+4$ versus $2+6$, or it might provide a more general characterization of the
exercise, e.g., \textit{single digit addition}.  In the latter case, exercise
are grouped by the \textit{skill} that must be applied to obtain a solution.
Although we will use the term skill in this article, others refer to the skill
as a \textit{knowledge component}, and the authors of DKT also use the term
\textit{concept}. Regardless, the important distinction for the purpose of
our work is between a label that indicates the particular exercise and a
label that indicates the general skill required to perform the exercise.
We refer to these two types of labels as \textit{exercise indexed} and
\textit{skill indexed}, respectively.

\subsection{Knowledge Tracing}

BKT models skill-specific performance, i.e., performance on a series of
exercises that all tap the same skill. A separate instantiation of BKT is made for
each skill, and a student's raw trial sequence is parsed into skill-specific
subsequences that preserve the relative ordering of exercises within a skill but
discard the ordering relationship of exercises across skills.  For a given skill
$\sigma$, BKT is trained using the data from each student $s$, $\{ X_{st} |
Y_{st}=\sigma \}$, where the relative trial order is preserved.  Because it
will become important for us to distinguish between absolute trial index and
the relative trial index within a skill, we use $t$ to denote the former and
use $i$ to denote the latter.

BKT is based on a theory of all-or-none human learning \cite{atkinson1972}
which postulates that the knowledge state of student $s$ following the $i$'th
exercise requiring a certain skill,
$K_{si}$, is binary: 1 if the skill has been mastered, 0 otherwise.
BKT, formalized as a hidden Markov model, infers $K_{si}$ from the
sequence of observed responses on trials $1 \ldots i$, $\{ X_{s1}, X_{s2},$
$\ldots,$ $X_{si}\}$. BKT is typically specified by four parameters: 
$P(K_{s0}=1)$, the probability that the student has mastered the skill
prior to solving the first exercise; $P(K_{s,i+1} = 1 ~|~ K_{si} = 0)$,
the transition probability from the not-mastered to mastered state;
$P(X_{si}=1 ~|~ K_{si}=0)$, the probability of correctly \emph{guessing} the
answer prior to skill mastery; and $P(X_{si}=0 ~|~ K_{si}=1)$, the
probability of answering incorrectly due to a \emph{slip} following skill
mastery.  Because BKT is typically used in modeling practice over brief
intervals, the model assumes no forgetting, i.e., $K$ cannot transition from 1
to 0.

BKT is a highly constrained, structured model. It assumes that
the student's knowledge state is binary, that predicting performance on 
an exercise requiring a given skill depends only on the student's binary 
knowledge state, and that the skill associated with each exercise is known in
advance. If correct, these assumptions allow the model to make strong
inferences. If incorrect, they limit the model's performance. The only way
to determine if model assumptions are correct is to construct an alternative
model that makes different assumptions and to determine whether the alternative
outperforms BKT.  DKT is exactly this alternative model, and its 
strong performance directs us to examine BKT's limitations. First, however,
we briefly describe DKT.

Rather than constructing a separate model for each skill, DKT models all skills
jointly. The input to the model is the complete sequence of exercise-performance
pairs, $\{ (X_{s1}, Y_{s1})$ \\$...  (X_{st}, Y_{st}) ...  (X_{sT}, Y_{sT})
\}$, presented one trial at a time.  As depicted in Figure~\ref{fig:nn},
DKT is a recurrent neural net which takes $(X_{st},Y_{st})$ as input and
predicts $X_{s,t+1}$ for each possible exercise label.  The model is trained and
evaluated based on the match between the actual and predicted $X_{s,t+1}$ for
the tested exercise ($Y_{s,t+1}$).  In addition to the input and output layers
representing the current trial and the next trial, respectively, the network
has a hidden layer with fully recurrent connections (i.e., each hidden unit
connects back to all other hidden units).  The hidden layer thus serves to
retain relevant aspects of the input history as they are useful for predicting
future performance.  The hidden state of the network can be conceived of as
embodying the student's knowledge state.  Piech et al.\ \cite{Piechetal2015}
used a particular type of hidden unit, called an LSTM (long short-term memory)
\cite{hochreiter1997long},
which is interesting because these hidden units behave very much like the BKT
latent knowledge state, $K_{si}$. To briefly explain LSTM, each hidden unit
acts like a memory element that can hold a bit of information. The unit is
triggered to turn on or off by events in the input or the state of other hidden
units, but when there is no specific trigger, the unit preserves its state,
very similar to the way that the latent state in BKT is sticky---once a skill
is learned it stays learned. With 200 LSTM hidden units---the number used in 
simulations reported in \cite{Piechetal2015}---and 50 skills, DKT has roughly 
250,000 free parameters (connection strengths).  Contrast this number with
the 200 free parameters required for embodying 50 different skills in BKT.
\begin{figure}
\centering
\includegraphics[height=1.7in]{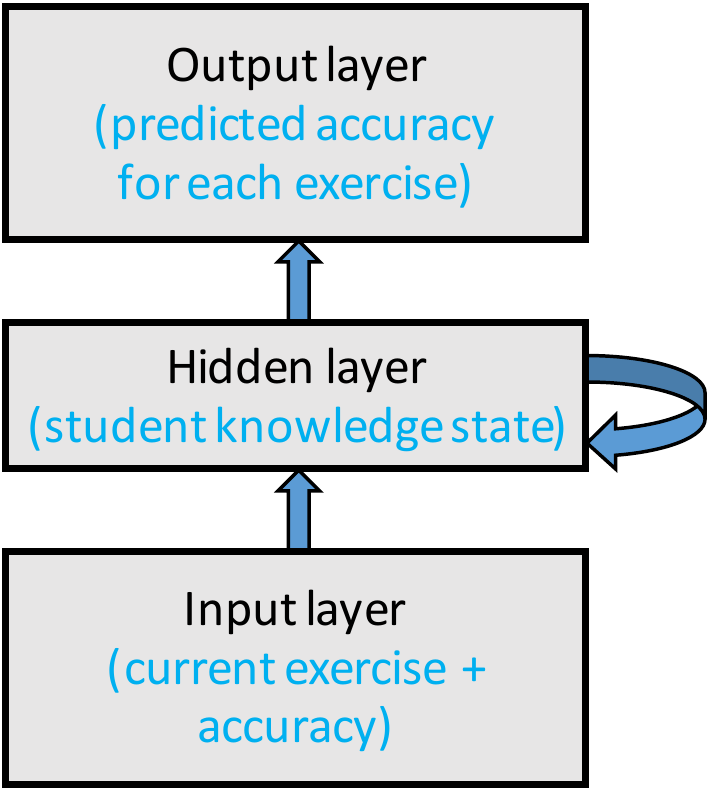}
\caption{Deep knowledge tracing (DKT) architecture. Each rectangle depicts
a set of processing units; each arrow depicts complete connectivity between
each unit in the source layer and each unit in the destination layer.
\label{fig:nn} }
\end{figure}

With its thousand-fold increase in flexibility, DKT is a very 
general architecture.  One can implement BKT-like dynamics in DKT with 
a particular, restricted set of connection strengths.  However, 
DKT clearly has the capacity to encode learning dynamics that are
outside the scope of BKT. This capacity is what allows DKT to discover
structure in the data that BKT misses.

\subsection{Where Does BKT Fall Short?}

In this section, we describe four regularities that we conjecture to be present
in the student-performance data. DKT is flexible enough that it has the
potential to discover these regularities, but the more constrained BKT model is
simply not crafted to exploit the regularities. In following sections, we
suggest means of extending BKT to exploit such regularities, and conduct
simulation studies to determine whether the enhanced BKT achieves performance
comparable to that of DKT.

\subsubsection{Recency Effects}

Human behavior is strongly recency driven. For example, when individuals
perform a choice task repeatedly, response latency can be predicted by an
exponentially decaying average of recent stimuli \cite{Jonesetal2013}.
Intuitively, one might expect to observe recency effects in student
performance. Consider, for example, a student's time varying engagement. If the
level of engagement varies slowly relative to the rate at which exercises are
being solved, a correlation would be induced in performance across local spans
of time. A student who performed poorly on the last trial because they were
distracted is likely to perform poorly on the current trial.  We conducted a
simple assessment of recency using the \assistments\ data set (the details of
this data set will be described shortly). Similarly to \cite{galyardt2015move}, 
we built an autoregressive model that
predicts performance on the current trial as an exponentially weighted average
of performance on past trials, with a decay half life of about 5 steps. 
We found that this single parameter model fit the \assistments\ data reliably
better than classic BKT. (We are not presenting details of this simulation
because we will evaluate a more rigorous variant of the idea in a following
section. Our goal here is to convince the reader that there is likely some
value to the notion of recency-weighted prediction.)

Recurrent neural networks tend to be more strongly influenced by recent events
in a sequence than more distal events \cite{Mozer1992}. Consequently, DKT is
well suited to exploiting recent performance in making predictions. In 
contrast, the generative model underlying BKT supposes that once a skill
is learned, performance will remain strong, and that a slip at time $t$
is independent of a slip at $t+1$. 

\subsubsection{Contextualized Trial Sequence}

The psychological literature on practice of multiple skills indicates that the
sequence in which an exercise is embedded influences learning and retention
(e.g., \cite{Rohreretal2014,Rohreretal2015}). For example, given three exercises
each of skills $A$ and $B$, presenting the exercises in the \textit{interleaved}
order $A_1$--$B_1$--$A_2$--$B_2$--$A_3$--$B_3$ yields superior performance relative to
presenting the exercises in the \textit{blocked} order $A_1$--$A_2$--$A_3$--$B_1$--$B_2$--$B_3$.
(Performance in this situation can be based on an immediate or delayed test.)

Because DKT is fed the entire sequence of exercises a student receives in the
order the student receives them, it can potentially infer the effect of exercise
order on learning. In contrast, because classic BKT separates exercises by skill,
preserving only the relative order of exercises within a skill,
the training sequence for BKT is the same regardless of whether the trial
order is blocked or interleaved.

\subsubsection{Inter-Skill Similarity}

Each exercise presented to a student has an associated label. In typical
applications of BKT---as well as two of the three simulations reported in Piech
et al.\ \cite{Piechetal2015}---the label indicates the skill required to solve
the problem. Any two such skills, $S_1$ and $S_2$, may vary in their degree of
relatedness. The stronger the relatedness, the more highly correlated one would
expect performance to be on exercises tapping the two skills, and the more
likely that the two skills will be learned simultaneously.

DKT has the capacity to encode inter-skill similarity.
If each hidden unit represents student knowledge state for a particular 
skill, then the hidden-to-hidden connections encode the degree of overlap.
In an extreme case, if two skills are highly similar, they can be
modeled by a single hidden knowledge state.  In contrast, classic BKT 
treats each skill as an independent modeling problem and thus can not 
discover or leverage inter-skill similarity.

DKT has the additional strength, as demonstrated by Piech et al., that it can
accommodate the absence of skill labels. If each label simply indexes a
specific exercise, DKT can discover interdependence between exercises in exactly
the same manner as it discovers interdependence between skills. In contrast,
BKT requires exercise labels to be skill indexed.

\subsubsection{Individual Variation in Ability}

Students vary in ability, as reflected in individual differences in mean
accuracy across trials and skills.  Individual variation might potentially be
used in a predictive manner: a student's accuracy on early trials in a sequence
might predict accuracy on later trials, regardless of the skills required to
solve exercises. We performed a simple verification of this hypothesis using the
\assistments\ data set.  In this data set, students study one skill at a time and
then move on to the next skill. We computed correlation between mean accuracy
of all trials on the first $n$ skills and the mean accuracy of all trials on
skill $n+1$, for all students and for $n \in \{1, ..., N-1\}$ where $N$ is the
number of skills a student studied. We obtained a correlation coefficient of
0.39: students who tend to do well on the early skills learned tend to do well
on later skills, regardless of the skills involved.

DKT is presented with a student's complete trial sequence. It can use a 
student's average accuracy up to trial $t$ to predict trial $t+1$.
Because BKT models each skill separately from the others, it does not have
the contextual information needed to estimate a student's average accuracy
or overall ability.

\section{Extending BKT}

In the previous section, we described four regularities that appear
to be present in the data and which we conjecture that DKT exploits but
which the classic BKT model cannot. In this section, we describe three 
extensions to BKT that would bring BKT on par with DKT with regard to these regularities.

\subsection{Forgetting}

To better capture recency effects, BKT can be augmented to allow
for forgetting of skills.  Forgetting corresponds to fitting a BKT parameter 
$F \equiv P(K_{s,i+1} = 0 ~|~ K_{si} = 1)$, the probability of transitioning from
a state of knowing to not knowing a skill. In standard BKT, $F=0$.

Without forgetting, once BKT infers that the student has learned, even a long
run of poorly performing trials cannot alter the inferred knowledge state.
However, with forgetting, the knowledge state can transition in either
direction, which allows the model to be more sensitive to the recent trials: A
run of unsuccessful trials is indicative of not knowing the skill, regardless
of what preceded the run.  Forgetting is not a new idea to BKT, and in fact was
included in the original psychological theory that underlies the notion of
binary knowledge state \cite{atkinson1972}. However, it has not typically been
incorporated into BKT.  When it has been included in BKT \cite{Qiuetal2011},
the motivation was to model forgetting from one day to the next, not forgetting
that can occur on a much shorter time scale.  

Incorporating forgetting can not only sensitize BKT to recent events but
can also contextualize trial sequences. To explain, consider
an exercise sequence such as $A_1$--$A_2$--$B_1$--$A_3$--$B_2$--$B_3$--$A_4$, where the labels
are instances of skills $A$ and $B$. Ordinary BKT discards the absolute
number of trials between two exercises of a given skill, but with forgetting,
we can count the number of intervening trials and treat each as an independent 
opportunity for forgetting to occur. Consequently, the probability of 
forgetting between $A_1$ and $A_2$ is $F$, but the probability of forgetting
between $A_2$ and $A_3$ is $1-(1-F)^2$ and between $A_3$ and $A_4$ is
$1-(1-F)^3$. Using forgetting, BKT can readily incorporate some information
about the absolute trial sequence, and therefore has more potential than
classic BKT to be sensitive to interspersed trials in the
exercise sequence.

\subsection{Skill Discovery}


To model interactions among skills, one might suppose that each skill has some
degree of influence on the learning of other skills, not unlike the connection
among hidden units in DKT. For BKT to allow for such interactions among skills,
the independent BKT models would need to be interconnected, using an
architecture such as a factorial hidden Markov model
\cite{GhahramaniJordan1996}. As an alternative to this somewhat complex
approach, we explored a simpler scheme in which different exercise labels
could be collapsed together to form a single skill. For example, consider an
exercise sequence such as $A_1$--$B_1$--$A_2$--$C_1$--$B_2$--$C_2$--$C_3$. If skills $A$ and $B$
are highly similar or overlapping, such that learning one predicts learning the
other, it would be more sensible to treat this sequence in a manner that groups
$A$ and $B$ into a single skill, and to train a single BKT instantiation on both
$A$ and $B$ trials. This approach can be used whether the exercise labels are
skill indices or exercise indices. (One of the data sets used by Piech
et al.\ \cite{Piechetal2015} to motivate DKT has exercise-indexed labels).

We recently proposed an inference procedure that automatically discovers the
cognitive skills needed to accurately model a given data set
\cite{LindseyKhajahMozer2014}. (A related procedure was independently proposed
in \cite{GonzalesBrenes2015}.) The approach couples BKT with a technique that
searches over partitions of the exercise labels to simultaneously (1) determine
which skill is required to correctly answer each exercise, and (2) model a
student's dynamical knowledge state for each skill.  Formally, the technique
assigns each exercise label to a latent skill such that a student's expected
accuracy on a sequence of same-skill exercises improves monotonically with
practice according to BKT. Rather than discarding the skills identified by
experts, our technique incorporates a nonparametric prior over the
exercise-skill assignments that is based on the expert-provided skills and a
weighted Chinese restaurant process \cite{ishwaran2003generalized}.

In the above illustration, our technique would group
$A$ and $B$ into one skill and $C$ into another. This procedure collapses like
skills (or like exercises), yielding better fits to the data by BKT.
Thus, the procedure performs a sort of \textit{skill discovery}.

\subsection{Incorporating Latent Student-Abilities}

To account for individual variation in student ability, we have extended BKT
\cite{LFKT,PALE} such that slip and guess probabilities are modulated by
a latent \textit{ability} parameter that is inferred from the data, much in the
spirit of item-response theory \cite{deboeck2004}. 
As we did in \cite{LFKT}, we assume that students with stronger abilities have lower slip and higher guess probabilities.
When the model
is presented with new students, the posterior predictive distribution on
abilities is used initially, but as responses from the new student are 
observed, uncertainty in the student's ability diminishes, yielding better
predictions for the student.

\section{Simulations}

\subsection{Data Sets}
Piech et al.\ \cite{Piechetal2015} studied three data sets. One of the data
sets, from Khan Academy, is not publicly available. Despite our requests and a
plea from one of the co-authors of the DKT paper, we were unable to obtain
permission from the data science team at Khan Academy to use the data set.  We
did investigate the other two data sets in Piech et al., which are as follows.

\textit{\assistments} is an electronic tutor that teaches and evaluates students in
grade-school math. The 2009-2010 ``skill builder'' data set is a large,
standard benchmark, available by searching the web for 
\textit{assistment-2009-2010-data}. We used the train/test split provided by
Piech et al., and following Piech et al., we discarded all students
who had only a single trial of data.

\textit{\synthetic} is a synthetic data set created by Piech et al.\ to model
virtual students learning virtual skills.  The training and test sets each
consist of 2000 virtual students performing the same sequence
of 50 exercises drawn from 5 skills. The exercise on trial $t$ is assumed to
have a difficulty characterized by $\delta_t$ and require a skill specified by
$\sigma_t$. The exercises are labeled by the \textit{identity of the exercise}, not by
the underlying skill, $\sigma_t$. The ability of a student, denoted, $\alpha_t$
varies over time according to a drift-diffusion process, generally increasing
with practice. The response correctness on trial $t$ is a Bernoulli draw with
probability specified by guessing-corrected item-response theory with
difficulty and ability parameters $\delta_t$ and $\alpha_t$. This data set
is challenging for BKT because the skill assignments, $\sigma_t$, are not provided
and must be inferred from the data. Without the skill assignments, BKT must
be used either with all exercises associated with a single skill or each 
exercise associated with its own skill. Either of these assumptions will
miss important structure in the data.  \synthetic\ is an interesting data set in
that the underlying generative model is neither a perfect match to DKT or
BKT (even with the enhancements we have described). The generative model
seems realistic in its assumption that knowledge state varies continuously.

We included two additional data sets in our simulations.
\textit{\spanish} is a data set of 182 middle-school students practicing 409 
Spanish exercises (translations and application of simple skills such
as verb conjugation) over the course of a 15-week semester, with
a total of 578,726 trials \cite{Lindsey2014}.
\textit{\statics} is from a college-level engineering statics
course with 189,297 trials and 333 students and 1,223 exercises
\cite{statics}, available from the PSLC DataShop web site 
\cite{datashop}.

\subsection{Methods}
We evaluated five variants of BKT\footnote{https://github.com/robert-lindsey/WCRP/tree/forgetting}, each of which incorporates a different 
subset of the extensions described in the previous section: a base
version that corresponds to the classic model and the model against which 
DKT was evaluated in \cite{Piechetal2015}, which we'll refer to simply
as \textit{BKT}; a version that incorporates forgetting (\textit{BKT+F}),
a version that incorporates skill discovery (\textit{BKT+S}), a version
that incorporates latent abilities (\textit{BKT+A}), and a version that 
incorporates all three of the extensions (\textit{BKT+FSA}).
We also built our own implementation of DKT with LSTM recurrent units\footnote{https://github.com/mmkhajah/dkt}.
(Piech et al. described the LSTM version as better performing, but
posted only the code for the standard recurrent neural net version.)
We verified that our implementation produced results comparable to those
reported in \cite{Piechetal2015} on \assistments\ and \synthetic. We
then also ran the model on \spanish\ and \statics.

For \assistments, \spanish, and \statics, we used a single train/test split.
The \assistments\ train/test split was identical to that used by Piech et al.\
For \synthetic, we used the 20 simulation sets provided by Piech et al.\ and 
averaged results across the 20 simulations.

Each model was evaluated on each domain's test data set, and the performance of
the model was quantified with a discriminability score, the \textit{area under
the ROC curve} or \textit{AUC}.  AUC is a measure ranging from .5, reflecting
no ability to discriminate correct from incorrect responses, to 1.0, reflecting
perfect discrimination. AUC is computed by obtaining a prediction on the test
set for each trial, across all skills, and then using the complete set of
predictions to form the ROC curve. Although Piech et al.\ \cite{Piechetal2015} 
do not describe the procedure they use to compute AUC for DKT, code they
have made available implements the procedure we describe, and not
the obvious alternative procedure in which ROC curves are computed on
a per-skill basis and then averaged to obtain an overall AUC.

\subsection{Results}
\definecolor{figr}{rgb}{1,.5,.5}
\definecolor{figy}{rgb}{.9,.83574,.45} 
\definecolor{figg}{rgb}{.6429,1,.5}
\definecolor{figc}{rgb}{.5,1,.7857}
\definecolor{figb}{rgb}{.5,.7857,1}
\definecolor{figm}{rgb}{.6429,.5,1}

Figure~\ref{fig:results} presents the results of our comparison of five
variants of BKT on the four data sets. We walk through the data sets from
left to right.
\begin{figure*}
\centering
\includegraphics[width=6.5in]{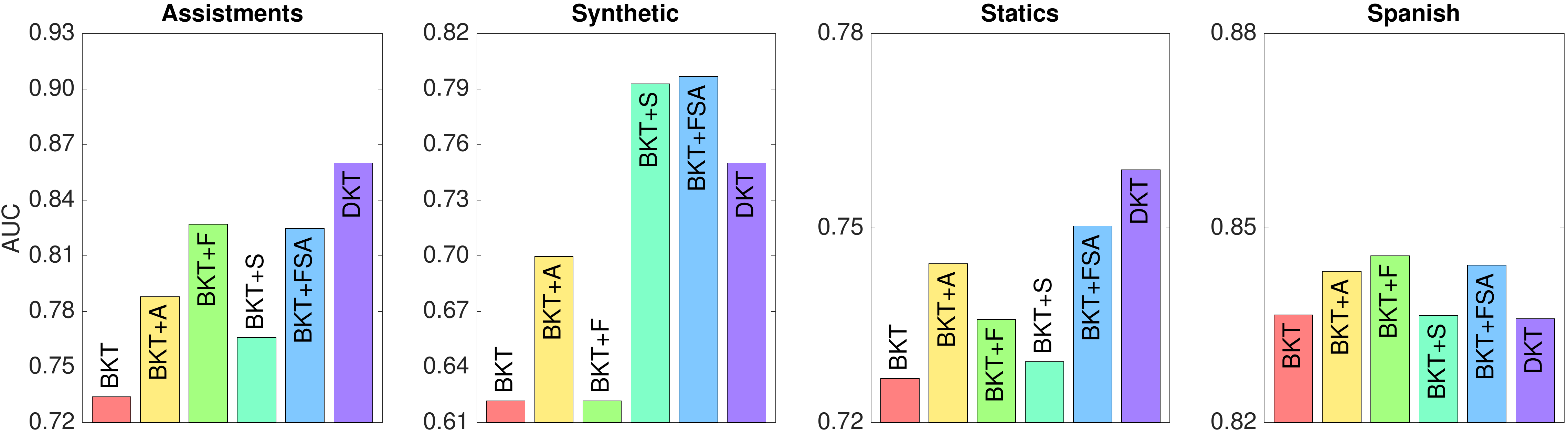}
\caption{A comparison of six models on four data sets. Model performance
on the test set is quantified by AUC, a measure of how well the model 
discriminates (predicts) correct and incorrect student responses. The models
are trained on one set of students and tested on another set. Note that the
AUC scale is different for each graph, but tic marks are always spaced by .03
units in AUC.
On
\assistments\ and \synthetic, DKT results are from Piech et al.\
\cite{Piechetal2015}; on
\statics\ and \spanish\, DKT results are from our own implementation.
\color{figr}BKT\color{black} = classic Bayesian knowledge tracing;
\color{figy}BKT+A\color{black} = BKT with inference of latent student abilities;
\color{figg}BKT+F\color{black} =
BKT with forgetting; 
\color{figc}BKT+S\color{black} = BKT with skill discovery; 
\color{figb}BKT+FSA\color{black} = BKT with all
three extensions; 
\color{figm}DKT\color{black} = deep knowledge tracing 
\label{fig:results}
}

\end{figure*}

On \assistments, classic BKT obtains an AUC of 0.73, better than the 0.67
reported for BKT by Piech et al. We are not sure why the scores do not match,
although 0.67 is close to the AUC score we obtain if we treat all exercises as
associated with a single skill or if we compute AUC on a per-skill basis and
then average.\footnote{Piech et al.\ cite Pardos and Heffernan
\protect\cite{pardos2011kt} as obtaining BKT's best reported performance on
\assistments---an AUC of 0.69. In \protect\cite{pardos2011kt}, the overall AUC
is computed by averaging the per-skill AUCs. This method yields a lower score
than the method used by Piech et al., for two reasons.  First, the Piech
procedure weighs all \textit{trials} equally, whereas the Pardos and Heffernan
procedure weighs all \textit{skills} equally.  With the latter procedure, the
overall AUC will be dinged if the model does poorly on a skill with just a few
trials, as we have observed to be the case with \assistments.  The latter
procedure also produces a lower overall AUC because it suppresses any lift due
to being able to predict the relative accuracy of different skills.  In
summary, it appears that inconsistent procedures may have been used to compute
performance of BKT versus DKT in \cite{Piechetal2015}, and the procedure for
BKT is biased to yield a lower score.} BKT+F obtains an AUC of 0.83, not quite
as good as the 0.86 value reported for DKT by Piech et al.\ Examining the
various enhancements to BKT, AUC is boosted both by incorporating forgetting
and by incorporating latent student abilities. We find it somewhat puzzling
that the combination of the two enhancements, embodied in BKT+FSA, does no
better than BKT+F or BKT+A, considering that the two enhancements tap different
properties of the data:  the student abilities help predict transfer from one
skill to the next, whereas forgetting facilitates prediction within a skill.

To summarize the comparison of BKT and DKT, 31.6\% of difference in performance
reported in \cite{Piechetal2015} appears to be due to the use of a biased
procedure for computing the AUC for BTK. Another 50.6\% of the difference in
performance reported vanishes if BKT is augmented to allow for forgetting.
We can further improve BKT if we allow the skill discovery algorithm to
operate with exercise labels that index individual exercises, as opposed
to labels that index the skill associated with each exercise.
With exercise-indexed labels, BKT+S and BKT+FSA both obtain an AUC of 0.90, 
beating DKT. However, given DKT's ability to perform skill discovery, we 
would not be surprised if it also achieved a similar level of performance 
when allowed to exploit exercise-indexed labels.

Turning to \synthetic, classic BKT obtains an AUC of 0.62, again significantly
better than the 0.54 reported by Piech et al. In our simulation, we treat each
exercise as having a distinct skill label, and thus BKT learns nothing more
than the mean performance level for a specific exercise. (Because the exercises
are presented in a fixed order, the exercise identity and the trial number are
confounded. Because performance tends to improve as trials advance in the
synthetic data, BKT is able to learn this
relationship.) It is possible here that Piech et al.\ treated all exercises as
associated with a single skill or that they used the biased procedure
for computing AUC; either of these explanations is consistent with
their reported AUC of 0.54.

Regarding the enhancements to BKT, adding student abilities (BKT+A) improves
prediction of \synthetic\ which is understandable given that the generative
process simulates students with abilities that vary slowly over time.  Adding
forgetting (BKT+F) does not help, consistent with the generative process which
assumes that knowledge level is on average increasing with practice; there is
no systematic forgetting in the student simulation.  Critical to this
simulation is skill induction: BKT+S and BKT+FSA achieve an AUC of 0.80, better
than the reported 0.75 for DKT in \cite{Piechetal2015}.

On \statics, each BKT extension obtains an improvement over classic BKT,
although the magnitude of the improvements are small.  The full model, BKT+FSA,
obtains an AUC of 0.75 and our implementation of DKT obtains a nearly identical
AUC of 0.76.  On \spanish, the BKT extensions obtain very little benefit. The
full model, BKT+FSA, obtains an AUC of 0.846 and again, DKT obtains a nearly
identical AUC of 0.836.  These two
sets of results indicate that for at least some data sets, classic BKT has no
glaring deficiencies. However, we note that BKT model accuracy can be improved 
if algorithms are considered that use exercise labels which are indexed by 
exercise and not by skill. For example, with \statics, performing skill 
discovery using exercise-indexed labels, 
\protect{\cite{Lindsey2014}} obtain an AUC of 0.81, much better than 
the score of 0.73 we report
here for BKT+S based on skill-indexed labels.

In summary, enhanced BKT appears to perform as well on average as DKT across
the four data sets.  Enhanced BKT outperforms DKT by 20.0\% (.05 AUC units) on
\synthetic\ and by 3.0\% (.01 AUC unit) on \spanish. Enhanced BKT underperforms
DKT by 8.3\% (.03 AUC units) on \assistments\ and by 3.5\% (.01 AUC unit) on
\statics.  These percentages are based on the difference of AUCs scaled by
by $\mathrm{AUC}_{\mathrm{DKT}}-0.5$, which takes into account the fact that 
an AUC of 0.5 indicates no discriminability.

\section{DISCUSSION}

Our goal in this article was to investigate the basis for the impressive
predictive advantage of deep knowledge tracing over Bayesian knowledge tracing.
We found some evidence that different procedures may have been used to evaluate
DKT and BKT in \cite{Piechetal2015}, leading to a bias against BKT. When we
replicated simulations of BKT reported in \cite{Piechetal2015}, we obtained
significantly better performance: an AUC of 0.73 versus 0.67 on \assistments,
and an AUC of 0.62 versus 0.54 on \synthetic.

However, even when the bias is eliminated, DKT obtains real performance gains
over BKT. To understand the basis for these gains, we hypothesized various
forms of regularity in the data which BKT is not able to exploit. We proposed
enhancements to BKT to allow it to exploit these regularities, and
we found that the enhanced BKT achieved a level of performance on average
indistinguishable from that of DKT over the four data sets tested.
The enhancements we explored are not
novel; they have previously been proposed and evaluated in the literature.
They include forgetting \cite{Qiuetal2011}, latent student abilities
\cite{LFKT,PALE,pardos2011kt}, and skill induction
\cite{Lindsey2014,GonzalesBrenes2015}.

We observe that different enhancements to BKT matter for different data sets.
For \assistments, incorporating forgetting is key; forgetting allows BKT to
capture recency effects. For \synthetic, incorporating skill discovery yielded
huge gains, as one would expect when the exercise-skill mapping is not known.
And for \statics, incorporating latent student abilities was relatively most
beneficial; these abilities enable the model to tease apart the capability
of a student and the intrinsic
difficulty of an exercise or skill.  Of the
three enhancements, forgetting and student abilities are computationally
inexpensive to implement, whereas skill discovery adds an extra layer of
computational complexity to inference.



The elegance of DKT is apparent when one considers the effort we have invested
to bring BKT to par with DKT.  DKT did not require its creators to analyze the
domain and determine sources of structure in the data. In contrast, our
approach to augmenting BKT required some domain expertise, a thoughtful
analysis of BKT's limitations, and distinct solutions to each limitation.  DKT
is a generic recurrent neural network model \cite{hochreiter1997long}, and it
has no constructs that are specialized to modeling learning and forgetting,
discovering skills, or inferring student abilities.  This flexibility makes DKT
robust on a variety of datasets with little prior analysis of the domains.
Although training recurrent networks is computationally intensive, tools exist
to exploit the parallel processing power in graphics processing units (GPUs),
which means that DKT can scale to large datasets. Classic BKT is inexpensive to
fit, although the variants we evaluated---particularly the model that
incorporates skill discovery---require computation-intensive MCMC methods that
have a distinct set of issues when it comes to parallelization.

DKT's advantages come at a price: interpretability.  DKT is massive
neural network model with tens of thousands of parameters which are
near-impossible to interpret. Although the creators of DKT did not have to
invest much up-front time analyzing their domain, they did have to invest
substantive effort to understand what the model had actually learned.  Our
proposed BKT extensions achieve predictive performance similar to DKT whilst
remaining interpretable: the model parameters (forgetting rate, student
ability, etc.) are psychologically meaningful.  When skill discovery is
incorporated into BKT, the result is clear: a partition of exercises into
skills. Reading out such a partitioning from DKT is challenging and only an
approximate representation of the knowledge in DKT.


Finally, we return to the question posed in the paper's title: How deep is
knowledge tracing?  Deep learning refers to the discovery of representations.
Our results suggest that representation discovery is not at the core of DKT's
success.  We base this argument on the fact that our enhancements to BKT bring
it to the performance level of DKT \textit{without} requiring any sort of
subsymbolic representation discovery.\footnote{Of course, the skill discovery
mechanism we incorporated certainly does regroup exercises to form skills, but
the form of this regrouping or partitioning is far more limited than the
typical transformations in a neural network to map from one level of
representation to another.} Representation discovery is clearly critical in
perceptual domains such as image or speech classification.  But the domain of
education and student learning is high level and abstract.  The input and
output elements of models are psychologically meaningful. The relevant internal
states of the learner have some psychological basis. The characterization of
exercises and skills can---to at least a partial extent---be expressed
symbolically.

Instead of attributing DKT's success to representation discovery, we attribute
DKT's success to its flexibility and generality in capturing statistical
regularities directly present in the inputs and outputs. As long as there are sufficient data
to constrain the model, DKT is more powerful than classic BKT.  BKT arose in a
simpler era, an era in which data and computation resources were precious. DKT
reveals the value of relaxing these constraints in the big data era.  But
despite the wild popularity of deep learning, there are many ways to relax the
constraints and build more powerful models other than creating a black box
predictive device with a vast interconnected tangle of connections and
parameters that are nearly impossible to interpret.

%
%
\end{abstract}

\section{Acknowledgments}
This research was supported by NSF grants SES-1461535, SBE-0542013, and
SMA-1041755.

%
\bibliographystyle{abbrv}
\bibliography{ref}  
%
%
\end{document}